\def\year{2022}\relax
\documentclass[letterpaper]{article} % DO NOT CHANGE THIS
\usepackage{aaai22}  % DO NOT CHANGE THIS
\usepackage{times}  % DO NOT CHANGE THIS
\usepackage{helvet}  % DO NOT CHANGE THIS
\usepackage{courier}  % DO NOT CHANGE THIS
\usepackage[hyphens]{url}  % DO NOT CHANGE THIS
\usepackage{graphicx} % DO NOT CHANGE THIS
\usepackage{natbib}  % DO NOT CHANGE THIS AND DO NOT ADD ANY OPTIONS TO IT
\usepackage{multirow}
\usepackage{caption} % DO NOT CHANGE THIS AND DO NOT ADD ANY OPTIONS TO IT
\DeclareCaptionStyle{ruled}{labelfont=normalfont,labelsep=colon,strut=off} % DO NOT CHANGE THIS
\usepackage{algorithm}
\usepackage{algorithmic}
\usepackage{xcolor}
\newcommand{\answerYes}[1]{\textcolor{blue}{#1}} 
 
\newcommand{\answerNA}[1]{\textcolor{gray}{#1}}

\usepackage{newfloat}
\usepackage{listings}
\floatstyle{ruled}
\newfloat{listing}{tb}{lst}{}
\floatname{listing}{Listing}

\newcommand{\anon}[1]{ANON.} % hide

\title{Coordinated Reply Attacks in Influence Operations: \\ Characterization and Detection}
\author{
    Manita Pote*\textsuperscript{\rm 1}\thanks{Corresponding author. Email: \texttt{potem@iu.edu}}, Tuğrulcan Elmas\textsuperscript{\rm 2}, Alessandro Flammini\textsuperscript{\rm 1}, 
    Filippo Menczer\textsuperscript{\rm 1}
}

\affiliations {
    \textsuperscript{\rm 1} Indiana University Bloomington\\
    \textsuperscript{\rm 2} University of Edinburgh\\
}

\begin{document}

\maketitle

\begin{abstract}
Coordinated reply attacks are a tactic observed in online influence operations and other coordinated campaigns to support or harass targeted individuals, or influence them or their followers.
Despite its potential to influence the public, past studies have yet to analyze or provide a methodology to detect this tactic.
In this study, we characterize coordinated reply attacks in the context of influence operations on Twitter. 
Our analysis reveals that the primary targets of these attacks are influential people such as journalists, news media, state officials, and politicians. 

We propose two supervised machine-learning models, one to classify tweets to determine whether they are targeted by a reply attack, and one to classify accounts that reply to a targeted tweet to determine whether they are part of a coordinated attack. 
The classifiers achieve AUC scores of 0.88 and 0.97, respectively. 
These results indicate that accounts involved in reply attacks can be detected, and the targeted accounts themselves can serve as sensors for influence operation detection. 
\end{abstract}

\section{Introduction}

Social media platforms are the primary environments in which civic engagement takes place. 
They play an important role in the exchange of ideas, discussion of political agendas, and development of political identities thanks to the ease with which one can access and consume information and build influence. 
However, social media platforms are also exploited by coordinated groups to purposefully distribute misleading information \cite{weedon_nuland_stamos_2017}, artificially amplify certain content \cite{elmas2022characterizing}, or interfere with elections \cite{ferrara2020characterizing, dni2017assessing, neudert2019sourcing}.
These types of social media exploitation are referred to as \textit{information operations} or \textit{influence operations} (IOs).

Influence operations are organized attempts to achieve a specific effect, such as manipulating public opinion, usually through coordinated tactics \cite{James2022}. 
IO tactics include public relations via advertising or paid digital influencers \cite{ong2018architects}; 
hashtag hijacking to distort trends and attract or distract the attention of mainstream media \cite{ong2018architects}; 
use of inauthentic and automated accounts to create the appearance of popularity \cite{elmas2023analyzing, woolley2019conclusion}; 
deletion of content violating terms of service to avoid detection by platforms \cite{Torres2022deletions}; 
troll accounts \cite{zannettou2019disinformation}; 
the spread of disinformation and propaganda \cite{woolley2019conclusion}; 
political memes \cite{rowett2018strategic, zannettou2020characterizing, ng2022coordinated};  
and `kompromat' strategies to influence political events \cite{woolley2019conclusion}. 

IOs can be state-sponsored, and originate domestically or in a foreign state \cite{bradshaw2017troops}.
A prime example of foreign-initiated campaign was the effort to interfere in the 2016 US Presidential Election by the Russian Internet Research Agency (IRA) \cite{senate_intelligence_committee_2019}. 
Reports on IOs from different countries like China, Brazil, and Nigeria show that such campaigns have emerged as a global threat \cite{bradshaw2017troops, woolley2019conclusion, bush:github}. 

Here we focus on \textit{coordinated reply attacks,} where a group of accounts work together to target specific individuals or entities by flooding their posts with replies. This can be done to overwhelm the target, push a particular narrative, or generate engagement. Coordinated reply attacks are actively employed in influence operations \cite{mathew2019trolling, bush:github}. 
Such a tactic has been used for harassment, as observed for example in a hate-speech campaign against Mehreen Faruqi, Australia's first female Muslim senator \cite{thomas_thompson_wanless_2020}; amplification by inauthentic accounts \cite{weedon_nuland_stamos_2017}; and spamming, trolling, and incitement. 

In this paper, we provide the first quantitative, large-scale study of coordinated reply attacks in influence operations reported by Twitter.\footnote{Although Twitter is now called X, we use the previous name because the data analyzed here predates the name change.} We explore the targeting patterns of IO actors employing this tactic and introduce methods to detect the targets of these attacks and the actors involved. 
We pose the following research questions: 
\begin{itemize}
    \item \textbf{RQ1}: Who are the targets of coordinated replies, and what specific topics characterize the tweets that attract such coordinated responses?
    \item \textbf{RQ2}: Among a set of tweets by potential targets, how can we identify those that receive coordinated replies? 
    \item \textbf{RQ3}: Given a set of targeted tweets, how can we detect the accounts that participate in coordinated replies?
\end{itemize}

We make the following contributions:
\begin{itemize}
    \item We find that primary targets of coordinated reply attacks are mostly influential people such as journalists, news media, state officials, and politicians. Most of the targets are attacked only once. The attacks for most of the targets are sporadic and tend to focus on specific contexts, such as politics. The attackers can originate from within the target's country or from foreign states.
    \item We present a classifier model to identify tweets that are targeted by coordinated replies. This model is generalizable to other contexts, as it does not use any features specific to IOs. It can also be developed into a tool for monitoring and safety. 
    \item We present a second model that performs well on the task of detecting accounts that are involved in reply attacks.  
\end{itemize}

\section{Related Work}

Given the dearth of prior research on coordinated reply attacks, we review the literature on IOs in general. 

\subsection{Characterization of Influence Operations}

Influence operations present novel challenges to content moderation on social media. A crucial initial step to address these challenges is to characterize how IO actors operate: their tactics, motivations, and engagement patterns.
\citet{mathew2019trolling} present different trolling techniques used in social media, from dogpiling to sock puppetry, along with interventions. 
\citet{zannettou2019disinformation} observe that Russian trolls displayed different behavior in the use of Twitter compared to random users. 
The same authors also find that Russian trolls on Twitter and Reddit were pro-Trump, while Iranian trolls were anti-Trump \cite{zannettou2019let}. 
The images shared by Russian trolls appeared in many popular social networks as well as mainstream and alternative news outlets, and focused on Russia, Ukraine, and the USA \cite{zannettou2020characterizing}. 
\citet{dutt2018senator} analyze the advertisements purchased by IRA accounts on Facebook and identify their changing campaign targets over time by performing clustering and semantic analysis. 
\citet{stewart2018examining} investigate the behavior of Russian trolls around the \#BlackLivesMatter movement and find that the trolls infiltrated both right- and left-leaning political communities to participate in both sides of the discussion. 
\citet{farkas2018} manually annotate IRA-linked tweets into 19 different categories to study whether IRA operations are consistent with classic propaganda models.
\citet{Merhi2021} find that the accounts involved in an IO in Turkey were resilient to large-scale shutdown. 
\citet{elmas2023misleading} discover that IO actors and other adversarial accounts often change their names and assume new identities. 
The \citet{sio} produced several reports describing influence campaigns and a range of tactics used by IO actors, including coordinated reply attacks \cite{bush:github}. 
While that work provides a qualitative description of the tactic, here we focus on methods to detect it. 

Coordinated reply attacks are also carried out by automated accounts; 
social bots have been reported to target influential users in an attempt to direct their attention toward fake news \cite{Shao18hoaxybots}.  
Financial rather than political incentives may be the drivers of such tactics, as in the case of cryptocurrency manipulation \cite{Yang2023Anatomy-AI-botnet}.  
The methods presented here are context-independent and therefore could be applied to these kinds of campaigns. 

\subsection{Detection of Influence Operations}

Many supervised machine-learning models have been proposed in the literature to detect IO actors, especially IRA trolls on Twitter, using deceptive linguistic cues~\cite{addawood2019linguistic} and behavioral and linguistic features~\cite{im2020still}. 
\citet{luceri2020detecting} propose an inverse reinforcement learning model for this task. 
\citet{alizadeh2020content-based} build a content-based classifier to detect tweets from troll accounts in Russia, China, and Venezuela IO campaigns.
Work from \citet{sharm2021hidden} uses a generative model to learn hidden group behavior to identify coordinated accounts. 
\citet{ezzeddine2023exposing} present an LSTM-based approach that identifies troll accounts based on behavioral cues. 
\citet{kong23Interval-censored} propose an interval-censored transformer Hawkes architecture to identify IO operators. 

Our work is similar to the above-mentioned efforts in the use of a supervised learning approach to identify coordinated accounts. However, we design features that leverage the targeting behaviors of the IO actors, specifically focusing on reply/comment engagements. Our method does not use any IO-specific features or sentiment cues, therefore it can be generalized to different social media platforms that have similar engagement functionalities.

Influence operations are one kind of coordinated campaign. 
A body of work has explored unsupervised methods to detect coordinated behaviors in general. \citet{pacheco2021uncovering} introduced a network-based framework for coordination detection. 
As campaigns use more than one tactic at a time, \citet{uyheng2022mapping} present a multi-view modularity clustering method. 
A Bayesian model by \citet{2024arXiv240106205H} leverages similarities in narrative and account characteristics. \citet{nwala2023language} propose a language framework that represents user actions and content as sequences of symbols to find coordinated accounts.
Unlike the above methods, we do not cluster accounts based on similar behaviors. We classify individual posts based on aggregate features of their replies, and individual accounts based on their metadata and reply activity. 

\section{Data Collection}

For the present study of coordinated reply attacks, we use 43 different state-sponsored IO datasets released by the Twitter Moderation Research Consortium from October 2018 to December 2021.\footnote{IO datasets were available at \url{transparency.x.com/en/reports/moderation-research} until summer 2024. An archival version of the site is available at \url{web.archive.org/web/20240829231920/https://transparency.x.com/en/reports/moderation-research}.} These datasets are archives of suspended accounts that Twitter claims to have been involved in foreign influence operations. Along with the account metadata, the datasets provide all the tweets generated by the accounts. 

Since according to Twitter the campaigns in these datasets are coordinated by a single entity, they provide us with ground truth for our study. Indeed, if we observe mass replies to a single target from multiple accounts labeled by Twitter as coordinated, we can establish that the coordinated reply attack tactic has been used. 

\subsection{Target Dataset}

First, we merge the datasets of all the IOs and keep all the replies by IO accounts to tweets by non-IO accounts. We refer to the latter accounts as \textit{targets} and to the replies by IO accounts as \textit{IO replies}. 
There are in total 17,873,714 IO replies from 44,425 IO accounts targeting 15,256,547 tweets by 1,763,084 distinct targets. 
From this data, we extract 15,016 targets and 96,041 tweets that received five or more direct replies from IO accounts. We assume these tweets have been targeted by coordinated reply attacks, and we label them as \textit{targeted tweets}. 
The threshold of five or more replies is arbitrary; a robustness analysis shows that the detection of targeted tweets does not seem to be affected by this parameter, as discussed later (Fig.~\ref{fig:threshold_performance}). 

The targeted tweets can still be publicly available at the time of our analysis, allowing us to collect all of their replies. 
Some of these replies may have originated from non-IO repliers, both before and after the IO accounts were taken down by Twitter. 
We refer to these replies as \textit{normal replies} and to their authors as \textit{normal repliers}. 
Since we only have direct replies by IO accounts, we only consider direct replies by normal repliers as well; replies to replies are discarded.
In addition to the metadata about the IO replies that are present in the initial data, we query the \texttt{/users/:id}, \texttt{/search/all}, and \texttt{/tweets/?ids=} endpoints of the Twitter API\footnote{\url{developer.twitter.com/en/docs/twitter-api/}} to collect metadata about the targets, the targeted tweets, their normal replies, and the normal repliers. 

Among the 15,016 targets, 5,041 were suspended, 3,992 could not be found (possibly deleted accounts), and 5,983 were alive at analysis time (2,031 verified and 3,952 non-verified accounts). Of the total 96,041 targeted tweets, 43,048 could not be found, which means these tweets could have originated from deleted or suspended accounts; 18,808 had unauthorized access; and 34,185 were accessible. For our influence operation case studies (\textbf{RQ1}), we consider this \textit{target dataset} of 34,185 tweets by 5,983 targets. 

\subsection{Classification Dataset}

\begin{figure}[t]
    \includegraphics[width=\columnwidth]{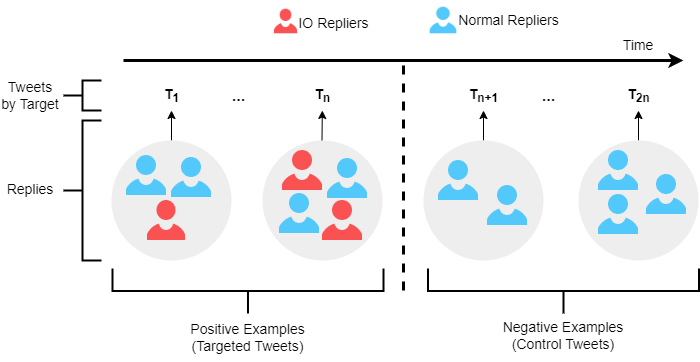}
    \caption{\small Data collection for the classifiers. The dashed line separated the last IO reply and the first successive tweet by the target.}
    \label{fig:classifier_dataset}
\end{figure}

To identify tweets targeted by coordinated replies (\textbf{RQ2}) and accounts that participate in such activity (\textbf{RQ3}), we consider targeted tweets as our positive examples. 
For corresponding negative examples, we collect \textit{control tweets} posted by the same targets after the last IO reply. This ensures that the tweets in our control data did not receive any coordinated replies by IO accounts. 
Fig.~\ref{fig:classifier_dataset} illustrates the data collection. 

As in the case of positive examples, we only retain control tweets with five or more replies.
In addition, to avoid bias due to the diverse activity of the targets, we collect from each target as many control tweets as targeted tweets.
Specifically, we select control tweets that were posted immediately after the last IO reply, subject to the five-reply minimum. 
In cases where we could not obtain as many control tweets as targeted tweets, we ensure a balanced dataset by keeping the most recent targeted tweets.

Similar to the targeted tweets, we fetch all the replies to the control tweets and all replier metadata. 
The resulting \textit{classification dataset} includes 3,866 targeted tweets and the same number of control tweets by 1,507 targets. 
There are in total 881,918 and 323,378 repliers in the positive and negative examples, respectively. These include IO and normal repliers. 
While the full classification dataset is used for \textbf{RQ2}, for \textbf{RQ3} we only use the positive examples (targeted tweets): 7,670 IO repliers and 874,248 normal repliers.

\section{RQ1: Targets and Topics}

In this section, we present an exploratory analysis of the targets of coordinated reply attacks and two case studies of specific campaigns where we can analyze the targets as well as the topics of their targeted tweets and other tactics employed in the campaigns. 

\begin{figure}
    \centering
    \includegraphics[width=\columnwidth]{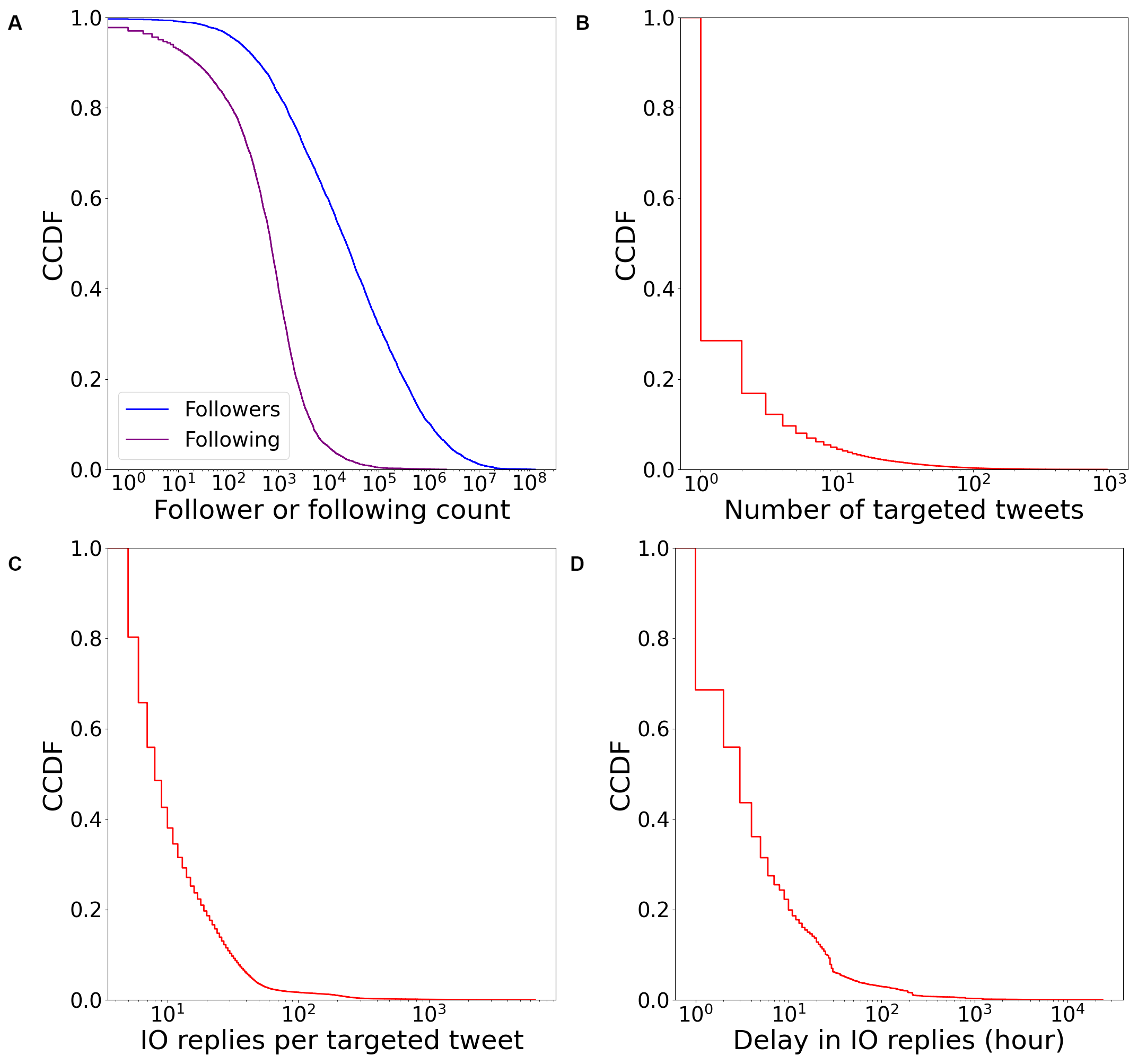}
    \caption{\small Complementary cumulative distribution functions (CCDF) of statistics describing target accounts and their tweets. 
    (A)~Numbers of followers and following (friends) of targets. (B)~Number of targeted tweets per target. (C)~Number of coordinated replies received by each targeted tweet. (D)~Time delay between targeted tweets and their coordinated replies.}
    \label{fig:exp_target}
\end{figure}

Exploratory analysis of target metadata (Fig.~\ref{fig:exp_target}a) shows that targets tend to have more followers (median 22,540) than following (median 707). 
This suggests that targets can be influential people. 
Reply attacks tend to be selective (Fig.~\ref{fig:exp_target}b): only a few tweets by each target were targeted (median one). 
The median number of coordinated replies received by targeted tweets was eight (Fig.~\ref{fig:exp_target}c).  
However, 54 of the targeted tweets received more than 1,000 replies. 
Coordinated replies tend to occur quickly after a targeted tweet, with a median delay of 3 hours (Fig.~\ref{fig:exp_target}d). 

To better understand what kinds of accounts were targeted, we annotated some target profiles with the corresponding professions or organization types and country of origin. 
We used manual annotation by checking each Twitter profile, description, the profession metadata indicated by the `briefcase icon,' and by searching Google for the accounts with more than a million followers. 
We grouped profession and organization types into broad categories, such as state officials, news media, and politicians. 
Accounts with insufficient information were labeled `Not Available.'

As the annotation process was time-consuming, we focused on two cases, namely two of the five campaigns with the most targets: Serbia (the top campaign with 1,175 targets) and Egypt (the fifth campaign with 372 targets).
In the next subsections, for each case, we report the top 10 target professions/types and countries.
We also inspected the targeted tweets to understand the context of the attacks. 
In preprocessing, we translated the targeted tweets into English and removed stop words and emojis.

\paragraph{Case Study: Serbia.}

\begin{figure}
    \centering
     \includegraphics[width=\columnwidth]{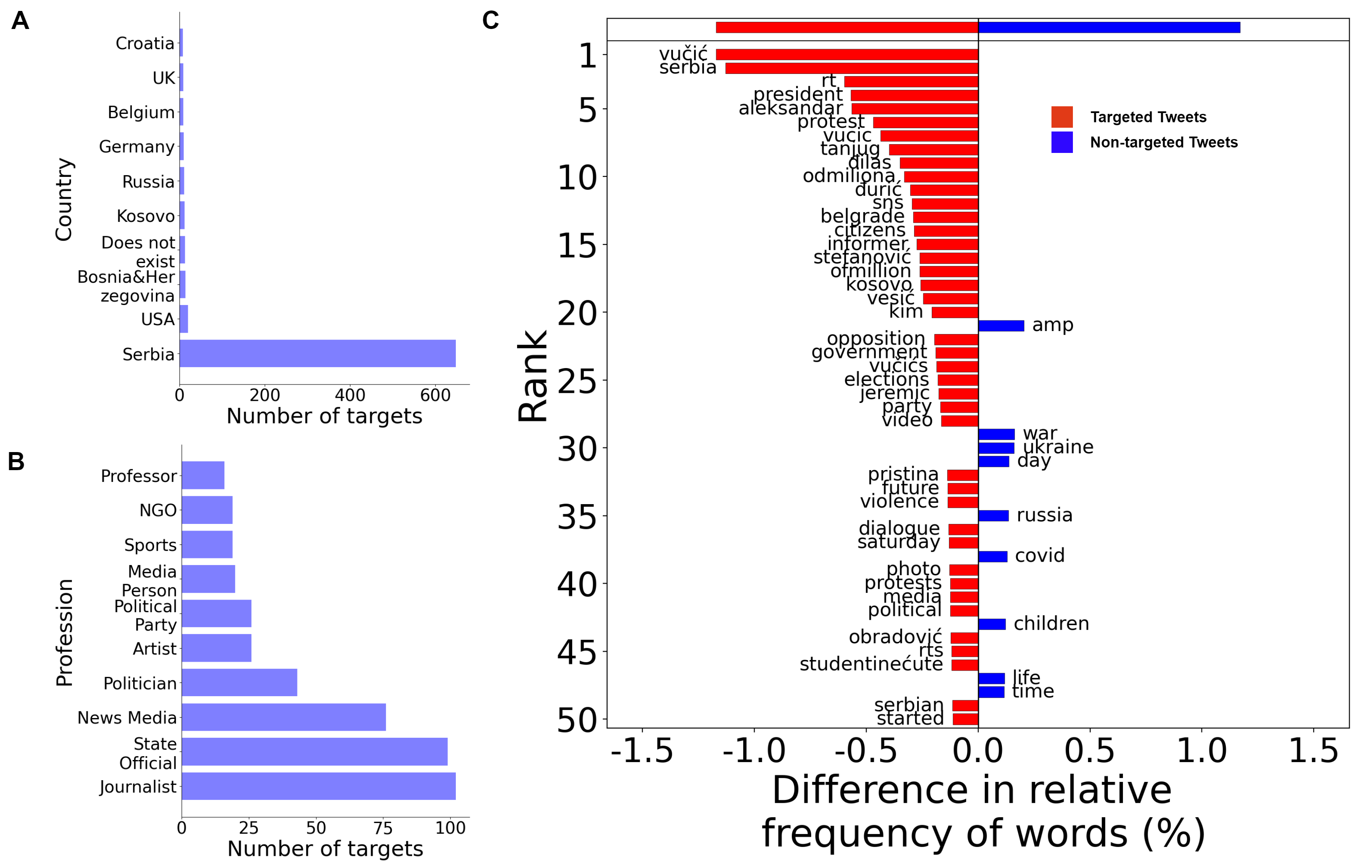}
    \caption{\small Characterization of the Serbia campaign. Distributions of (A) countries and (B) professions of the targets. (C) Wordshift graph comparing the most frequent words in targeted and non-targeted tweets.}
    \label{fig:serbia_campaign}
\end{figure}

The majority of accounts targeted by the Serbia campaign, approximately 648, were from Serbia itself, with the remaining coming from the Balkan region (Fig.~\ref{fig:serbia_campaign}a).  
This suggests that the campaign focused its efforts on influencing public opinion within Serbia.
Fig.~\ref{fig:serbia_campaign}b reveals that the coordinated reply attacks primarily targeted journalists (102), state officials (99), news media organizations (76), and politicians (43). 
A wordshift graph \cite{wordshift} highlighting the most prominent terms in the targeted tweets (Fig.~\ref{fig:serbia_campaign}c) shows that the campaign focused on President Vucic, the Serbian Progressive Party (SNS), the 2017 election, the ``1 out of 5 Millions" protest, and the Serbia-Kosovo diplomatic crisis. 
These findings are consistent with analysis by \citet{bush:github}, who reported that the primary objective of IO actors involved in the Serbia campaign was to rally support for President Alexander Vucic and his party, the SNS. 
This was achieved by promoting the popularity and visibility of Vucic and the SNS through retweeting their content and replying to other accounts with supportive messages. 
The IO accounts also targeted opponent political parties with derisive tweets and attempted to discredit them by flooding their posts with negative comments. 
This tactic aimed to create a public perception that the opposition was unpopular.

\paragraph{Case Study: Egypt.}

\begin{figure}[t]
    \centering
    \includegraphics[width=\columnwidth]{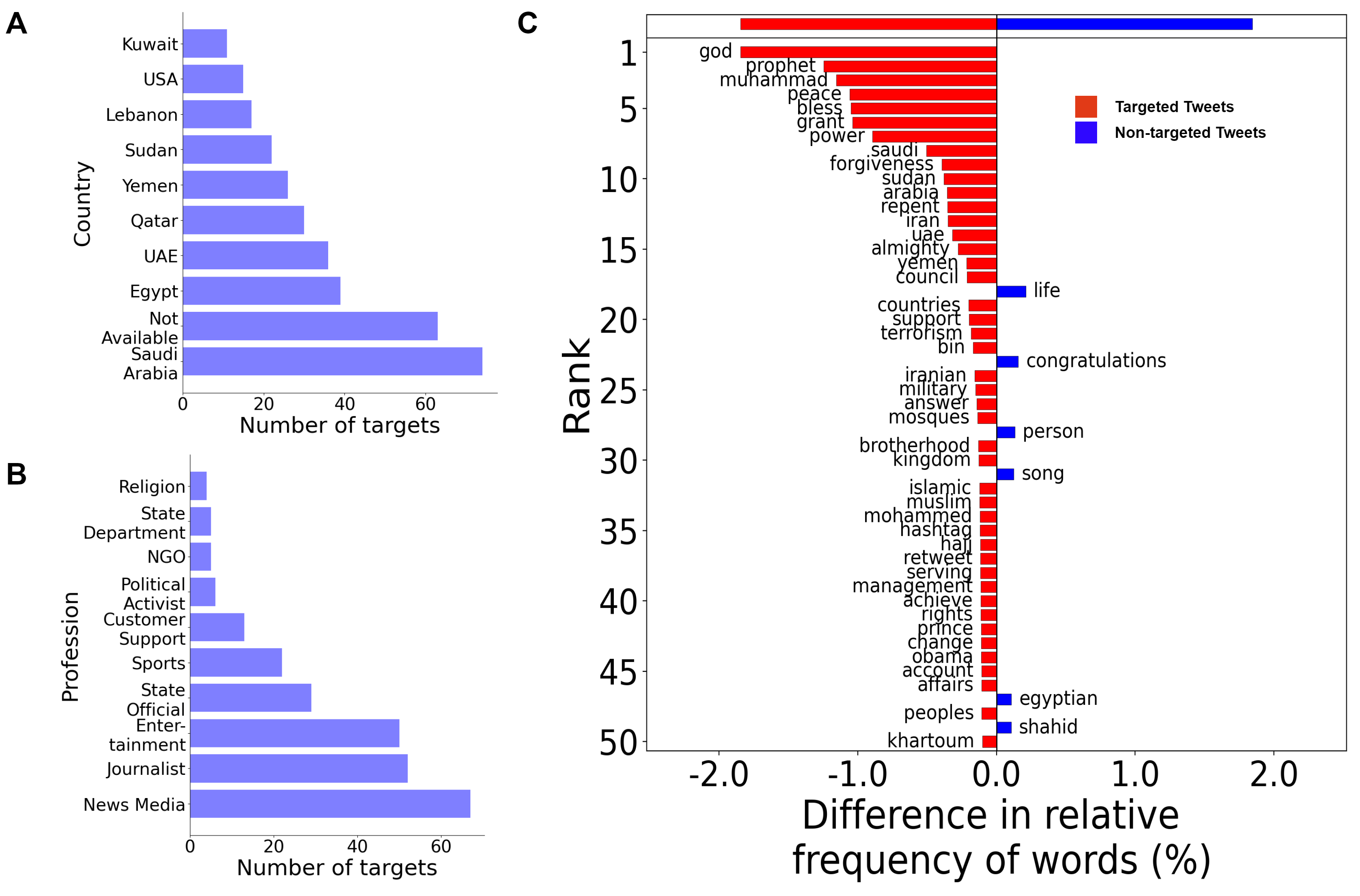}
    \caption{\small Characterization of the Egypt campaign. Distributions of (A) countries and (B) professions of the targets. (C) Wordshift graph comparing the most frequent words in targeted and non-targeted tweets.}
    \label{fig:egpyt_campaign}
\end{figure}

Fig.~\ref{fig:egpyt_campaign}a shows that the majority of accounts targeted by the Egypt campaign were from multiple Middle East and North Africa countries, primarily Saudi Arabia (74 targets), Egypt (39), UAE (36), Qatar (30), and Yemen (26).
This suggests a potential interstate attack. 
News media organization (67), journalists (52), and state officials (29) were again the main targets of the coordinated replies (Fig.~\ref{fig:egpyt_campaign}b). 
The analysis of common terms in the targeted tweets (Fig.~\ref{fig:egpyt_campaign}c) and manual inspection reveal that the Egypt campaign primarily focused on religious themes, terrorism, and current affairs like the Iran Nuclear deal (2018), Yemen's Houthi movement, Sudan's military coup, and the Muslim Brotherhood. 
These observations are consistent with a report by \citet{egypt_report}, describing an IO activity orchestrated by Egypt and the UAE, supporting the Saudi and Egyptian governments and criticizing Qatar, Turkey, Yemen, Iran. 

Both case studies indicate that influential people like journalists, news media, state officials, and politicians, are the primary targets of coordinated reply attacks.
These targets can be from different countries than the campaign's country of origin. 
The topics of the targeted tweets depend on the current affairs of the specific geographic region or country.

\section{RQ2: Tweet Classification}

Identifying the tweets that receive inauthentic coordinated replies is a necessary first step toward the detection of both the targets and the perpetrators of a coordinated attack. 
To address this challenge, we propose a campaign-independent classifier for identifying IO-targeted tweets. The same methodology could also be generalized to platforms other than Twitter. 

\subsection{Classifier Features}
\label{sec:rq1features}

 \begin{figure*}
    \centering
   \includegraphics[width=1.4\columnwidth]{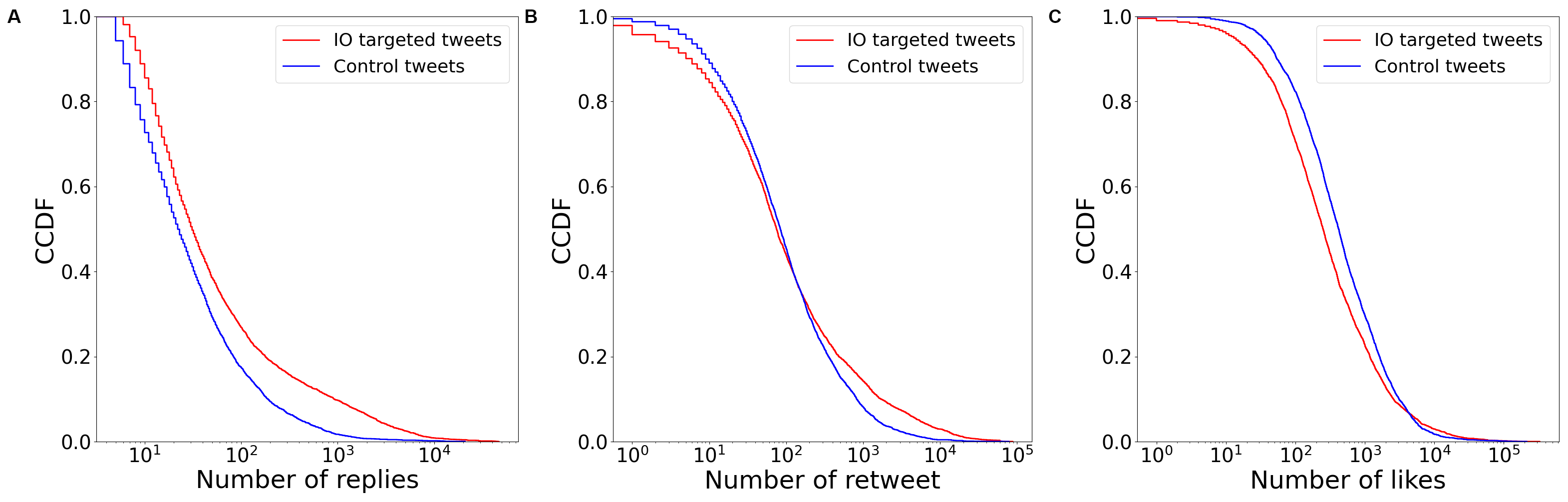}
    \caption{\small Engagement received by targeted and control tweets. (A)~Replies, (B)~retweets, and (C)~likes.}
    \label{fig:exp_tweet}
\end{figure*}

The tweet classifier leverages several features extracted from tweets and from the replies they receive. Let us first focus on tweet-level features, specifically tweet engagement. 
We find a few key differences between the engagement metrics of IO-targeted vs.~control tweets. 
As illustrated in Fig.~\ref{fig:exp_tweet}a, 
IO-targeted tweets receive more replies (median 31 vs.~22 for control tweets). 
On the other hand, control tweets receive slightly more retweets (median 84 vs.~75 for IO-targeted tweets, Fig.~\ref{fig:exp_tweet}b) and more likes (median 420 vs.~250, Fig.~\ref{fig:exp_tweet}c). 
This suggests that organic engagement generated more positive interactions and sharing, while inauthentic activity mainly focused on manipulating conversations through replies. 
Based on these observations, we use three tweet-level features: \texttt{reply\_count}, \texttt{retweet\_count}, and \texttt{like\_count}.

\begin{table}
    \centering
    \caption{\small Reply-level attributes used to generate features for the tweet classifier.}
    \small
    \begin{tabular}{ll}
        \hline 
       \textbf{Set} & \textbf{Attributes} \\
        \hline \hline
        \multirow{3}{*}{Engagement} 
        & \texttt{like\_count}\\
        & \texttt{retweet\_count}\\
        & \texttt{reply\_count}\\
        \hline
         \multirow{3}{*}{Entities} & \texttt{mention\_count} \\
        & \texttt{hashtag\_count} \\
        & \texttt{url\_count} \\
        \hline 
        Delay & \texttt{reply\_time\_diff} \\
         \hline
        Similarity & \texttt{cosine} \\
        \hline
    \end{tabular}
    \label{tab:tweet_features}
\end{table}

Next, let us consider reply-level features. These are based on eight attributes, listed in Table~\ref{tab:tweet_features}. 
Engagement and entity attributes are defined for each reply. 
The delay is also defined, for each reply, as the difference between the timestamps of the tweet and the reply.  
The similarity is designed to capture the presence of similar narratives in replies, a common characteristic of inauthentic engagement. 
To this end, we first generate vector embeddings for the replies using the LaBSE model~\cite{feng2020language}, which supports 109 languages. 
The \texttt{cosine} attribute is then computed for each pair of replies to the same tweet as the cosine similarity between the corresponding vectors.

Since targeted tweets can have many replies, this procedure yields many attribute values that must be aggregated to obtain a set of features for each tweet. 
In the case of engagement, entities, and delay attributes, we have one value per reply.  
For the similarity attribute, we have one value per a pair of replies. 
In all cases, we aggregate these values to obtain a single distribution of attribute values for each tweet. 
From these distributions we compute the following 12 summary statistic features: range, 25/50/75 quartiles, inter-quartile range, minimum, maximum, mean, standard deviation, skewness, kurtosis, and entropy. 
Since we do this for each of eight attributes, the total number of reply-level features used in the classifier is $8 \times 12 = 96$. Including the three tweet-level features, the classifier uses a total of 99 features.

\subsection{Results}

We compare different machine learning models: Logistic Regression, Random Forest, AdaBoost, Decision Tree, and Naive Bayes. 
Prior to training, we standardize the input features via z-scores. 
We conduct 10-fold cross-validation to mitigate over-fitting of the training data and report on the mean precision, recall, and F1 values across folds along with AUC in Table~\ref{tab:result_tweet_classifier}. 
Precision, recall, and F1 depend on a threshold to transform the model score into a binary classification label. We tune the threshold to maximize the mean F1 across folds. 
In the following, we focus on Random Forest (with 100 estimator trees), which yields the best scores overall.

\begin{table*}
    \centering
    \caption{\small Results of different algorithms in the tweet classification task. We present standard errors rounded to the second decimal point.}
    \small
    \begin{tabular}{lcccc}
    \hline
    \textbf{Classifier} & \textbf{Prec.} & \textbf{Rec.} & \textbf{F1} &  \textbf{AUC}\\
    \hline
    Logistic Regression & 0.65 $\pm$ 0.00 & 0.86 $\pm$ 0.00 & 0.74 $\pm$ 0.00 & 0.80 $\pm$ 0.00 \\
    Random Forest & 0.73 $\pm$ 0.00 & 0.87 $\pm$ 0.00 & 0.80 $\pm$ 0.00 & 0.88 $\pm$ 0.00\\
    AdaBoost & 0.64 $\pm$ 0.00 & 0.89 $\pm$ 0.00 & 0.74 $\pm$ 0.00 & 0.81 $\pm$ 0.00 \\
    Decision Tree & 0.52 $\pm$ 0.01 & 0.95 $\pm$ 0.02 & 0.66 $\pm$ 0.00 & 0.69 $\pm$ 0.00 \\
    Naive Bayes & 0.49 $\pm$ 0.00 & 1.00 $\pm$ 0.00 & 0.66 $\pm$ 0.00 & 0.68 $\pm$ 0.00 \\
    \hline
    \end{tabular}
    \label{tab:result_tweet_classifier}
\end{table*}

\begin{table} 
    \centering
    \caption{\small Contributions of different tweet-level features and reply-level feature sets to the Random Forest tweet classifier. The last row (using all features) corresponds to the results in Table~\ref{tab:result_tweet_classifier}.}
    \small
    \begin{tabular}{lcccc}
    \hline
    \textbf{Features set} & \textbf{Prec.} & \textbf{Rec.} & \textbf{F1} &  \textbf{AUC}\\
    \hline
    \texttt{reply\_count}     & 0.5 & 0.99 & 0.67 & 0.59 \\
    \texttt{retweet\_count}   & 0.49 & 1 & 0.66 & 0.54 \\
    \texttt{like\_count}      & 0.49 & 1 & 0.66 & 0.52 \\
    Engagement                              & 0.69 & 0.86 & 0.77 & 0.84\\
    Entities                                & 0.52 & 0.95 & 0.67 & 0.65\\
    Delay                                   & 0.51 & 0.96 & 0.67 & 0.66\\
    Similarity                              & 0.54 & 0.96 & 0.69 & 0.68\\
    \hline
    All features                            & 0.73 & 0.87 & 0.80 & 0.88\\
    \hline
    \end{tabular}
    \label{tab:diff_feat_performance}
\end{table}

\begin{figure}[t]
    \centerline{\includegraphics[width=.8\columnwidth]{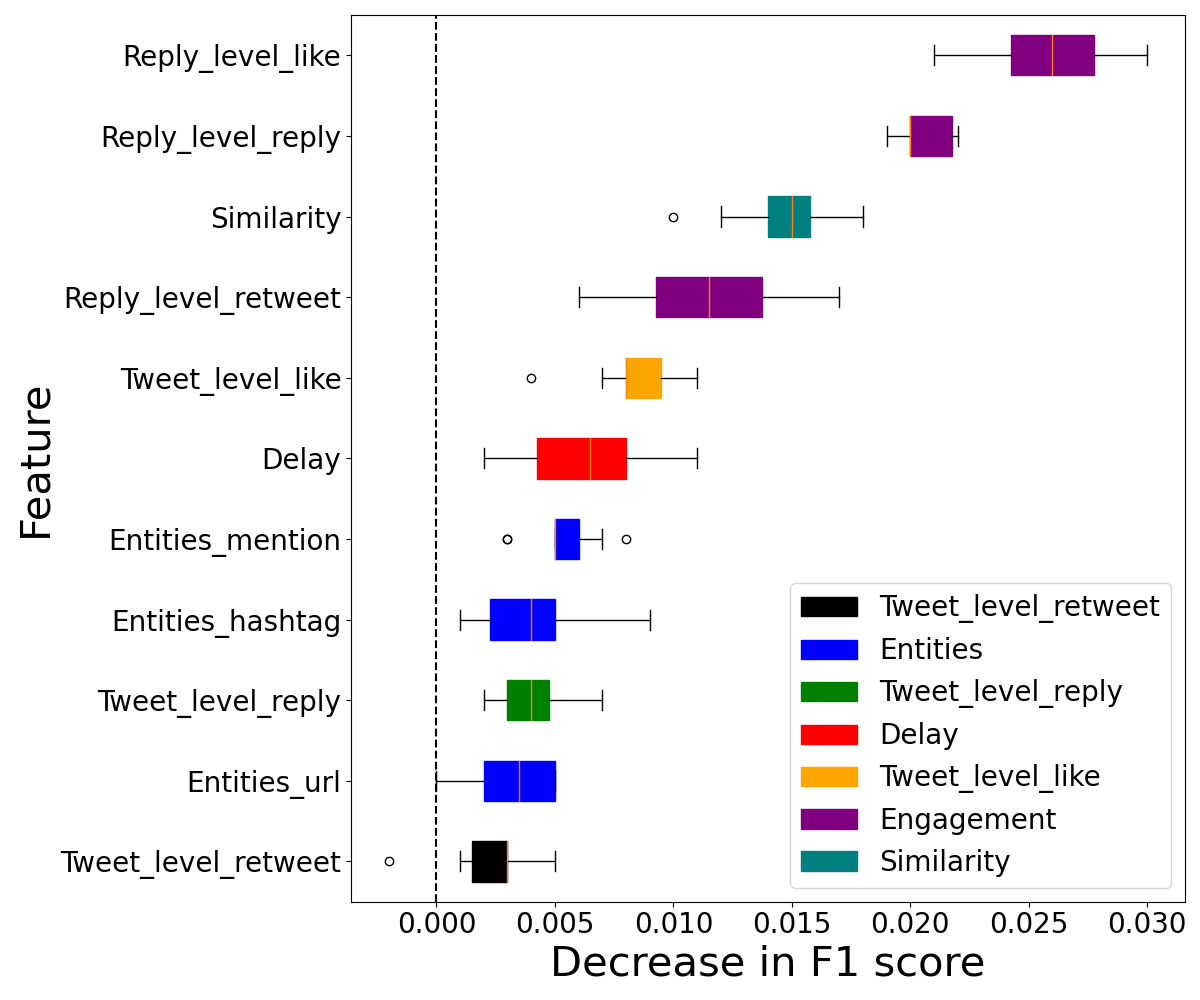}}
    \caption{\small Permutation feature importance for tweet classifier. We report the median (orange line), 50\% confidence interval (box), and 99.3\% confidence interval (whiskers) of the drop in F1 score when each feature/attribute is shuffled. Boxes with the same color indicate attributes in the same feature set. Larger values indicate higher importance.}
    \label{fig:feature_import_tweet}
\end{figure}

To study the contributions of different features, we followed two approaches.
First, we trained and tested Random Forest on individual tweet-level features and reply-level feature sets.
The results using 10-fold cross-validation are given in Table~\ref{tab:diff_feat_performance}.
Second, we performed a permutation feature importance test, which measures the importance of features by computing the loss in accuracy when the values of those features are shuffled (permuted). %for each tweet-level feature and reply-level attribute.
To simplify the analysis, for each reply-level attribute we shuffled all the corresponding features rather than each feature individually. 
For example, for the \texttt{like\_count} engagement attribute, we shuffled all 12 summary statistics features at once.
We repeated this test 10 times and recorded the drop in mean F1 score from 10-fold cross-validation for each iteration.
The distribution of these drop values is given in Fig.~\ref{fig:feature_import_tweet}. 

Both approaches consistently show that reply-level engagement features are the most important. A classifier using only those features achieves F1=0.77 and AUC=0.84 (Table~\ref{tab:diff_feat_performance}), and removing those features causes significant drops in F1 (Fig.~\ref{fig:feature_import_tweet}).
In our classification dataset, the majority of targets are from the Serbia campaign. IO accounts in this campaign were not intended to generate engagement with other Twitter users; instead, they primarily boosted retweet and reply counts for other IO accounts to artificially amplify Vucic and his allies on Twitter \cite{bush:github}. 
In fact, we find that replies to targeted tweets from IO accounts have a higher mean retweet count than replies from normal repliers (0.63 vs 0.31) and also a higher mean like count (1.28 vs 0.81) and mean reply count (0.20 vs 0.13). 
However, our data does not allow us to determine if such engagement was mostly driven by IO accounts or organic. 

Since reply-level engagement may be affected by the popularity of the targeted tweets, it is legitimate to ask whether tweet-level engagement features would provide sufficient signals to discriminate between targeted and control tweets. 
However, Table \ref{tab:diff_feat_performance} indicates that tweet-level reply, like, and retweet counts do not provide very informative signals for tweet classification.
To further explore this question, let us measure the correlation between tweet-level features (\texttt{reply\_count}, \texttt{retweet\_count}, and \texttt{like\_count}) and the corresponding reply-level engagement counts.
As each original tweet can have many replies, there are many more replies than original tweets. 
We therefore calculate the mean correlation between pairs of tweet/reply engagement features across 10 random samples of replies matching the number of original tweets. 
The correlations are all very small (around 0.001), confirming that reply engagement is not a mere reflection of tweet popularity.

\begin{figure}[t]
    \centerline{\includegraphics[width=.6\columnwidth]{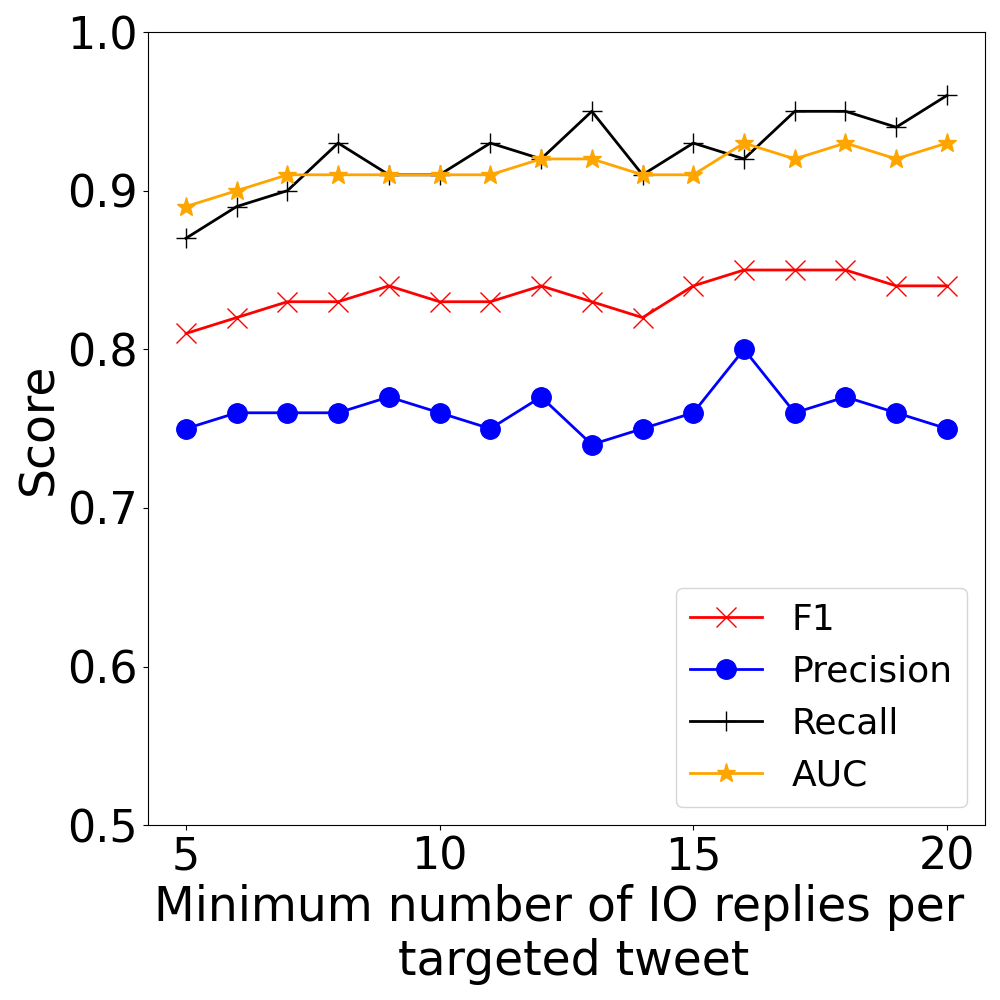}}
    \caption{\small Scores of tweet classifiers based on different thresholds for the number of IO replies received by targeted tweets.}
    \label{fig:threshold_performance}
\end{figure}

We previously defined \textit{targeted tweets} as those that receive five or more replies from IO accounts. 
Let us test the robustness of our classifier with respect to this definition by considering a range of threshold values between five and 20 replies from IO accounts. 
This filters down the set of targeted tweets and corresponding control tweets. 
We follow the same procedure described above to construct the classification dataset, extract the features, and train and evaluate the classifier.  
Fig.~\ref{fig:threshold_performance} reports the mean precision, recall, F1, and AUC from 10-fold cross-validation. 
While we observe slight increases as the criterion for defining targeted tweets becomes more stringent, the results appear to be robust with respect to this parameter.

\begin{table}
    \centering
    \caption{\small F1 scores obtained from same-campaign (diagonal entries, in bold) and cross-campaign evaluations of the tweet classifier. RS=Serbia, SA=Saudi Arabia, TR=Turkey, EG=Egypt. SA/EG/AE is a campaign involving three countries.}
    \resizebox{\columnwidth}{!}{\begin{tabular}{l|cccccc}
    \hline
    \multirow{2}{*}{\textbf{Train}} & \multicolumn{6}{c}{\textbf{Test}} \\
    \cline{2-7}
                  & RS       & SA & TR & EG   & SA/EG/AE & Other \\
    \hline
      RS      & \textbf{0.85}         & 0.54         & 0.61   & 0.56    & 0.52        & 0.65 \\
    SA  & 0.55         & \textbf{0.76}          & 0.53  & 0.63    & 0.74        & 0.68\\
    TR        & 0.61          & 0.60        & \textbf{0.74}  & 0.63    & 0.65         & 0.68\\
    EG         & 0.43          & 0.37         & 0.36  & \textbf{0.65}    & 0.38        & 0.39\\
    SA/EG/AE    & 0.47         & 0.58         & 0.57  & 0.60    & \textbf{0.73}        & 0.48\\
    Other         &  0.62          & 0.59        & 0.63  & 0.56    & 0.52        & \textbf{0 .74} \\
    \hline
    \end{tabular}}
    \label{tab:cross_camp_performance}
\end{table}

Next, let us evaluate the generality of the classifier by testing how well a model trained on one campaign performs when tested on other campaigns.
First, we split the classification dataset into six subsets: one for each of the top five campaigns, based on the number of targeted tweets, and one with data aggregated from the remaining campaigns.
Second, we train campaign-specific models on each of these datasets, as in the original tweet classification setup. 
Finally, we evaluate the models on test data from each dataset.  
In the diagonal of Table \ref{tab:cross_camp_performance} we report F1 values when the model trained on one campaign is tested on the same campaign (mean across 10-fold cross-validation). The off-diagonal F1 values are obtained when the model trained on all data from one campaign (optimized to maximize F1) is tested on other campaigns.
As expected, the models perform better when trained and tested on the same campaign.
However, models can generalize, with F1 drops that depend on the specific campaigns.
This suggests that at least some commonalities exist across coordinated reply campaigns.

\section{RQ3: Replier Classification}

Once we identify potentially targeted tweets, we can attempt to detect, among the accounts that reply to them, those that are engaged in coordinated activity.
Distinguishing authentic replies from inauthentic ones poses a non-trivial challenge, given that inauthentic accounts attempt to create an impression of authenticity. 
For this task, we train a supervised replier classifier using the targeted tweet dataset. 

\subsection{Classifier Features}

\begin{figure}
    \centering
    \includegraphics[width=\columnwidth]{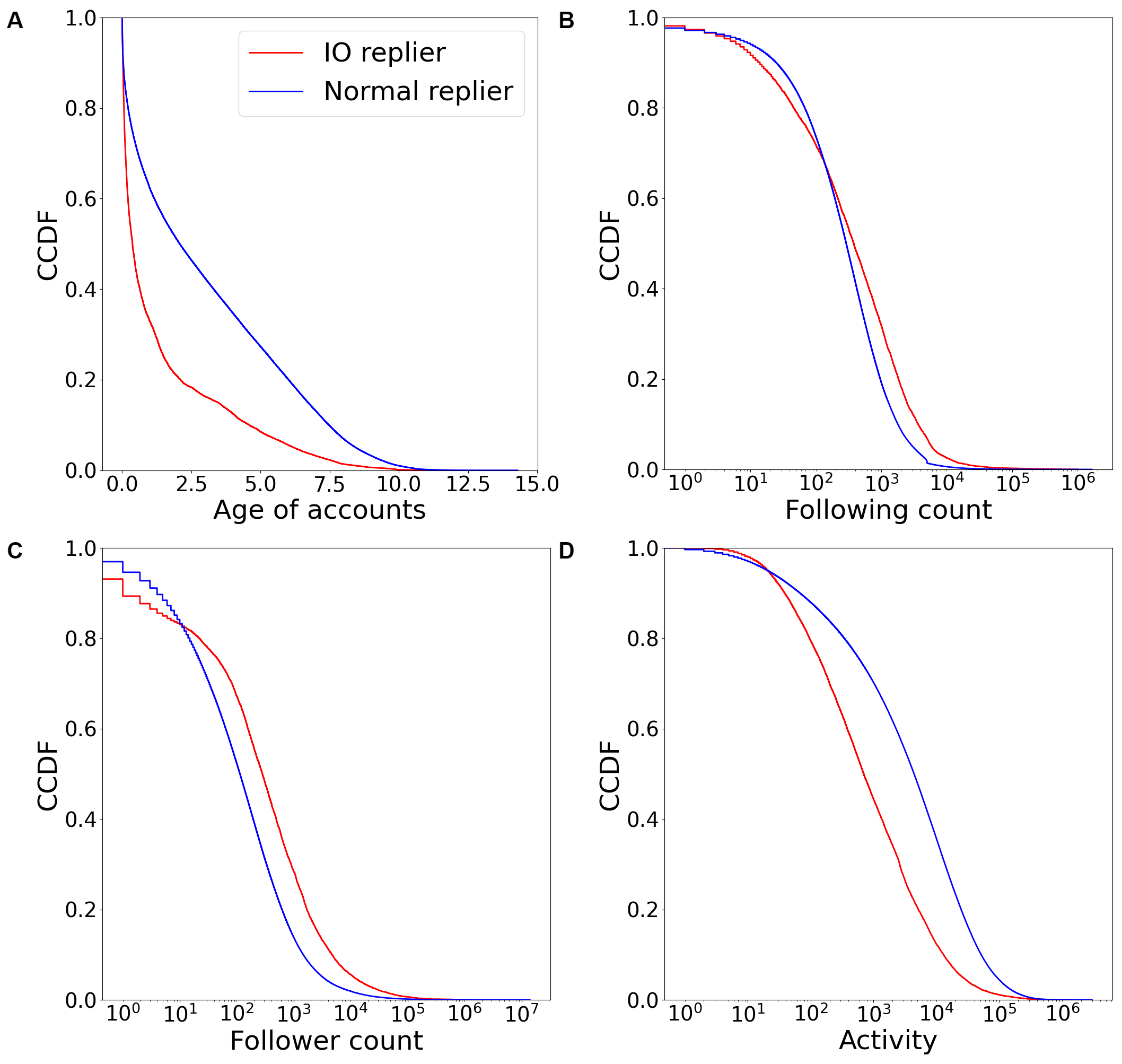}
    \caption{\small Differences between complementary cumulative distributions of IO and normal replier metadata: (A) age, (B) following count, (C) follower count, and (D) activity, as measured by the sum of the numbers of original tweets, replies, quotes, and retweets.}
    \label{fig:IO_normal_replier_characteristic_diff}
\end{figure}

We engineer features for each replier from their profile metadata and their replies to the targeted tweets. 
Starting with profile metadata, we calculate the age of repliers by subtracting the account creation date from the date of the last reply by the account. 
As illustrated in Fig.~\ref{fig:IO_normal_replier_characteristic_diff}(A), most IO repliers are relatively new accounts, with a median age of 0.37 years compared to 2.08 years for normal repliers. 
Despite their relatively young age, IO repliers display a higher median number of followers (282) and followings (380) compared to normal repliers, whose medians are 114 and 292, respectively (Figure \ref{fig:IO_normal_replier_characteristic_diff}(B, C)). 
However, IO repliers exhibit lower activity levels (original tweets + replies + quotes + retweets), with a median of 699 compared to 4406 for normal repliers (Figure \ref{fig:IO_normal_replier_characteristic_diff}(D)).

To leverage these key differences between IO and normal repliers, we create four features specific to profile metadata: \texttt{age}, \texttt{follower\_rate}, \texttt{following\_rate}, and \texttt{activity\_rate}. 
Since the numbers of followers/following and the activity are correlated with an account's age, we normalize the rate features by the age of the account. 

Each replier may be involved in one or more replies to multiple targeted tweets.
Therefore, we create a number of features that summarize the characteristics of the set of replies generated by each replier, including replies to multiple targeted tweets. 
These features are based on eight reply attributes, which we organize into four sets, just like those listed in  Table~\ref{tab:tweet_features}. 
The only criterion that distinguishes how these features are calculated in the tweet versus the replier classification task is the reply set --- all replies to a tweet in the former case and all replies by a replier in the latter. 

Given a set of replies, the replier classifier features are calculated as for the tweet classifier, with two exceptions. 
First, the delay of each reply is computed with respect to the timestamp of the targeted tweet to which the reply was directed.
Second, cosine similarity $s$ for replier $i$ is calculated for each pair $(r^t_i, r^t_j)$ where $r^t_i$ is a reply by $i$ to a targeted tweet $t$ and $r^t_j$ is a reply by a different user $j$ to the same targeted tweet $t$. We obtain a distribution of these similarities $\cup_{t \in T(i)} \cup_{j \in J(t)} s(r^t_i, r^t_j)$ across the set $J(t)$ of other users who reply to $t$ and then across the set $T(i)$ of targeted tweets that receive a reply from $i$. 

From the distribution of each attribute, we compute nine summary statistic features: range, 25/50/75 quartiles, inter-quartile range, maximum, minimum, mean, and entropy. 
We do not calculate standard deviation, skewness, and kurtosis because they are not defined for many repliers who are involved in a single reply to a single targeted tweet.

We end up with four profile metadata features and $8 \times 9 = 72$ reply-level features, for a total of 76 features.

\subsection{Results}

The targeted tweet dataset is highly imbalanced with 0.8\% IO repliers (7,670 vs.~874,248 normal repliers).
Such an imbalance leads to poor classification, which can be addressed in two ways. 
First, we downsampled the normal repliers by creating 10 different balanced datasets. Each includes all the IO repliers and an equal number (7,670) of normal repliers, sampled without replacement. 
We train and test the model on each balanced dataset using 10-fold cross-validation and report the average performance score. 
As a second approach, we over sampled the IO repliers by splitting the data into train and test sets, then replicating the minority class. Replication occurs only in the training data, to avoid data leakage. We run 10-fold cross-validation on the resulting dataset. 
The first approach might eliminate some potential false positives --- normal repliers with similar reply behavior --- potentially making the task easier.  
In the second approach, the model is tested on data that still maintains the class imbalance, potentially overfitting the training data. This approach is also more expensive due to the large dataset. 
Given these complementary disadvantages, below we report on both methods. 

\begin{table*}
    \centering
    \caption{\small Results of different algorithms in the replier classification task. Top: downsampling of the majority class (normal repliers). Bottom: oversampling of the minority class (IO repliers). We present standard errors rounded to the second decimal point.} 
    \small
    \begin{tabular}{lcccc}
    \hline
    \textbf{Downsampling} & \textbf{Prec.} & \textbf{Rec.} & \textbf{F1} &  \textbf{AUC}\\
    \hline
    Logistic Regression & 0.89 $\pm$ 0.00 & 0.88 $\pm$ 0.00 & 0.88 $\pm$ 0.00 & 0.93 $\pm$ 0.00\\
    Random Forest & 0.93 $\pm$ 0.00 & 0.92 $\pm$ 0.00  & 0.92 $\pm$ 0.00 & 0.97 $\pm$ 0.00\\
    AdaBoost      & 0.90 $\pm$ 0.00 & 0.90 $\pm$ 0.00 & 0.90 $\pm$ 0.00 & 0.96 $\pm$ 0.00 \\
    Decision Tree & 0.88 $\pm$ 0.00 & 0.88 $\pm$ 0.00 & 0.88 $\pm$ 0.00 & 0.88 $\pm$ 0.00\\
    Naive Bayes & 0.62 $\pm$ 0.02 & 0.86 $\pm$ 0.02     & 0.68 $\pm$ 0.00 & 0.87 $\pm$ 0.00 \\
    \hline
    \hline
    \textbf{Oversampling} & \textbf{Precision} & \textbf{Recall} & \textbf{F1} &  \textbf{AUC}\\
    \hline
    Logistic Regression & 0.27 $\pm$ 0.00     & 0.48 $\pm$ 0.01 & 0.35 $\pm$ 0.00 & 0.94 $\pm$ 0.00 \\
    Random Forest       & 0.70 $\pm$ 0.00     & 0.72 $\pm$ 0.01 & 0.71 $\pm$ 0.00 & 0.96 $\pm$ 0.00\\
    AdaBoost            & 0.47 $\pm$ 0.02    & 0.54 $\pm$ 0.02  & 0.50 $\pm$ 0.01 & 0.96 $\pm$ 0.00 \\
    Decision Tree       & 0.55 $\pm$ 0.01   & 0.51 $\pm$ 0.01  & 0.53 $\pm$ 0.01  & 0.75 $\pm$ 0.01 \\
    Naive Bayes         & 0.06 $\pm$ 0.00  & 0.50 $\pm$ 0.03    & 0.10 $\pm$ 0.00 & 0.86 $\pm$ 0.00\\
    \hline
    \end{tabular}
    \label{tab:result_replier_classifier}
\end{table*}

We standardize the features with z-scores and report the mean performance metrics obtained by different machine learning models: Logistic Regression, Random Forest, AdaBoost, Decision Tree, and Naive Bayes.
As in the tweet classifier, we tune the threshold to maximize the mean F1 across folds. 
Table~\ref{tab:result_replier_classifier} shows that all classifiers perform better with downsampling, and Random Forest (with 100 estimator trees) performs the best with both downsampling and oversampling. 
Therefore, let us focus on this model --- Random Forest trained with downsampling --- for further analysis.

\begin{table}[t]
    \centering
    \caption{\small Contributions of different profile features and reply-level feature sets to the Random Forest replier classifier. The last row corresponds to the results in Table~\ref{tab:result_replier_classifier} (top).}
    %\begin{tabular}{|p{0.2\columnwidth}|p{0.1\columnwidth}|p{0.1\columnwidth}|p{0.1\columnwidth}|}
    \small
    \begin{tabular}{lcccc}
    \hline
    \textbf{Features set} & \textbf{Prec.} & \textbf{Rec.} & \textbf{F1}   &  \textbf{AUC}\\
    \hline    
    \texttt{activity\_rate}         & 0.60               & 0.63           & 0.61                & 0.65 \\
    \texttt{following\_rate}         & 0.54               & 0.52               & 0.53              & 0.56 \\
    \texttt{follower\_rate}          &  0.55              & 0.51               &0.53                  & 0.56 \\
     \texttt{age}             & 0.58               & 0.66           & 0.61                & 0.66 \\
    Delay                 & 0.57               & 0.60            & 0.58                & 0.62\\
    Engagement            & 0.57               & 0.57            & 0.53                & 0.63\\
    Entities              & 0.63               & 0.50           & 0.53                & 0.63\\
    Similarity            & 0.85      & 0.84  & 0.84     & 0.92\\
    \hline
    All features         & 0.93               & 0.92            & 0.92               & 0.97\\
    \hline
    \end{tabular}
    \label{tab:replier_feat_performance}
\end{table}

\begin{figure}
    \centerline{\includegraphics[width=0.8\columnwidth]{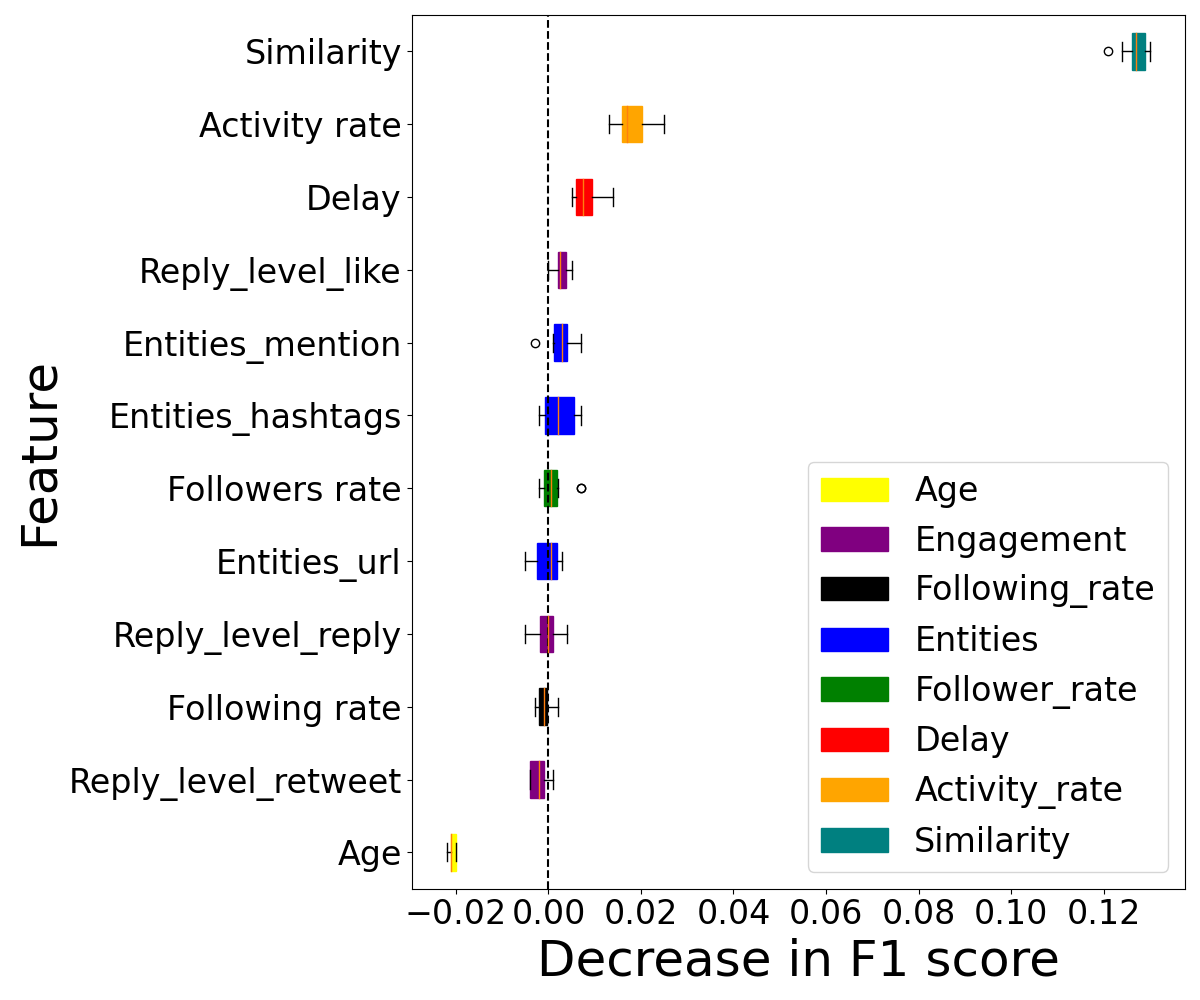}}
    \caption{\small Permutation feature importance for replier classifier. We report the median (orange line), 50\% confidence interval (box), and 99.3\% confidence interval (whiskers) of the drop in F1 score when each feature/attribute is shuffled. Boxes with the same color indicate attributes in the same feature set. Larger values indicate higher importance.}
    \label{fig:feature_replier}
\end{figure}

\begin{figure}[t]
    \centerline{
     \includegraphics[width=\columnwidth]{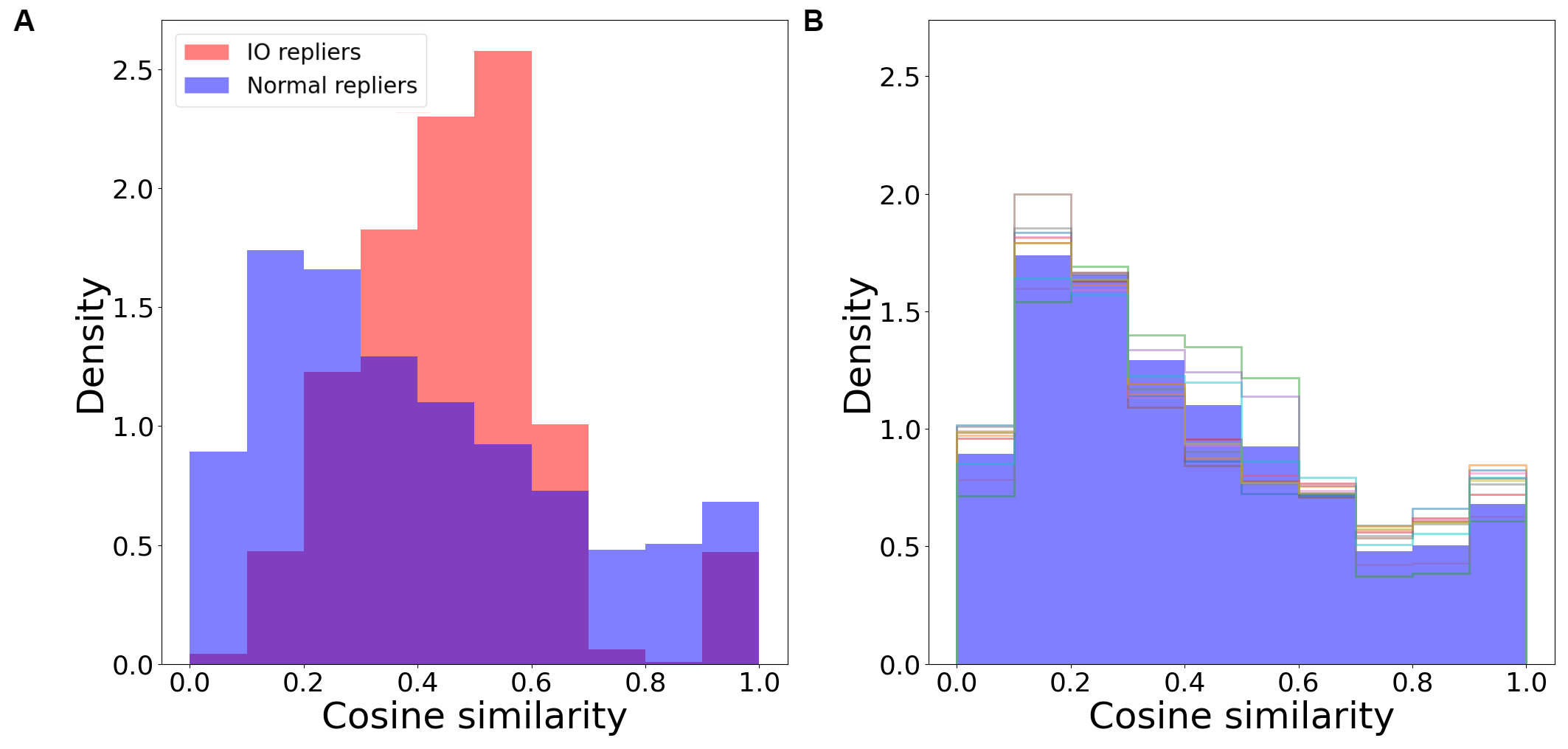}
    }
    \caption{\small (A) Distributions of cosine similarity attributes for replies by IO and normal repliers. (B) Similarity distributions for normal repliers (blue, same as in panel (A)) and the 10 different samples (different colored outlines).}
    \label{fig:cosine_similarity_repliers}
\end{figure}

To test the contribution of each feature to the replier classifier, we follow the same procedure as for the tweet classifier. However, here we report the averages across the 10 balanced dataset. 
Table~\ref{tab:replier_feat_performance} reports on mean 10-fold cross-validation scores for Random Forest trained and tested on each profile metadata feature and reply feature set. 
Fig.~\ref{fig:feature_replier} reports on the results of a permutation feature importance test. 
Both analyses consistently shows that the similarity among replies is the most important feature. 
To help interpret this finding, Fig.~\ref{fig:cosine_similarity_repliers}(A) compares the distributions of similarity attributes for replies by IO versus normal repliers. Replies by IO repliers are more similar to other replies to the same tweets, compared to those by normal repliers. This is a pattern that the classifier can exploit. 
We also observe in Fig.~\ref{fig:cosine_similarity_repliers}(B) that the downsampling process does not bias the similarity distributions. 

\begin{figure}
    \centering
    \includegraphics[width=0.6\columnwidth]{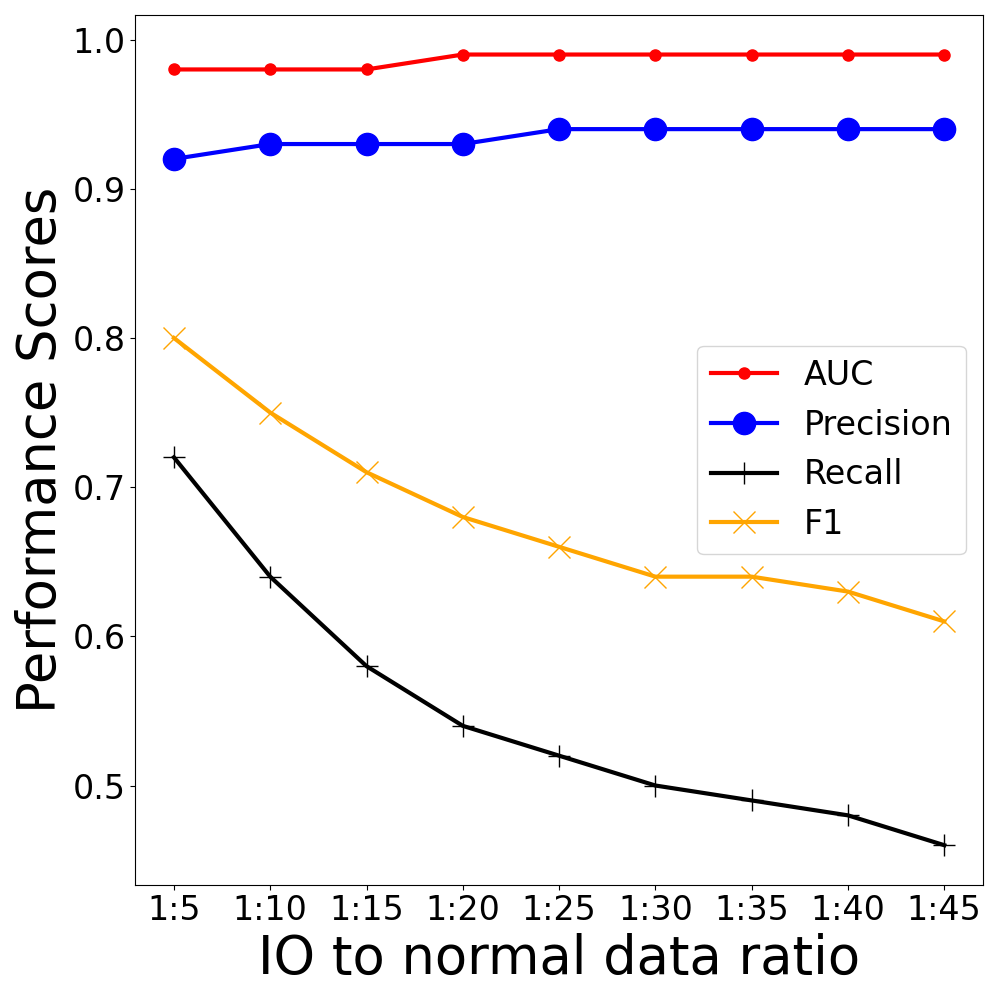}
    \caption{\small Replier classification scores for different data imbalance ratios.}
    \label{fig:datasize_performance_result}
\end{figure}

So far, we have tested the replier classifier on balanced datasets. 
In more realistic scenarios, the data may be imbalanced with different ratios of coordinated and organic repliers. 
To test whether the classifier can generalize to such scenarios, we train and test with 10-fold cross-validation using different positive to negative data ratios ranging from 1:5 to 1:45.
Fig.~\ref{fig:datasize_performance_result} shows that precision and AUC are robust to class imbalance, whereas recall (and consequently F1) drops as the class imbalance increases.
While this result indicates that balancing the classes affects recall, we also found that training on the original imbalanced dataset leads to a high false-positive rate and deteriorated precision.  

Next, we test the generalizability of the replier classifier by training and testing the model across different campaigns.
Similarly to the tweet classifier, we prepare six campaign datasets: five selected based on the highest number of IO repliers and one by aggregating the remaining campaigns. 
In the diagonal of Table \ref{tab:replier_cross_camp_performance} we report F1 values when the model trained on one campaign is tested on the same campaign (mean across 10-fold cross-validation and 10 different balanced datasets). The off-diagonal F1 values are obtained when the model trained on one of the balanced datasets (selected to maximize F1) is tested on other campaigns.
We observe that the campaign-specific models tend to perform well both on their own campaigns and across campaigns. 
The Serbia campaign is an exception, where we observe a sizable deterioration in cross-campaign evaluation.  
These results suggest that the features of the replier classifier are relevant across different campaigns.

\begin{table}
    \centering
    \caption{\small F1 scores obtained from same-campaign (diagonal entries, in bold) and cross-campaign evaluations of the replier classifier. HN=Honduras; see Table~\ref{tab:cross_camp_performance} for other country codes.}
    \small
    \begin{tabular}{l|cccccc}
    \hline
    \multirow{2}{*}{\textbf{Train}} & \multicolumn{6}{c}{\textbf{Test}} \\
    \cline{2-7}
                   &   SA & RS  & TR & EG & HN & Other \\
    \hline
     SA &  \textbf{0.96}  & 0.79          & 0.86          & 0.86         & 0.55           & 0.90 \\
    RS         &  0.53           & \textbf{0.89} & 0.76          & 0.53          & 0.48          & 0.49\\
    TR         & 0.90            & 0.91          & \textbf{0.92} & 0.84          & 0.84          & 0.85\\
    EG          & 0.92            & 0.90          & 0.88          & \textbf{0.92} & 0.74          & 0.90\\
    HN       & 0.93            & 0.97          & 0.98          & 0.89          & \textbf{0.95} & 0.91\\
    Other      & 0.95            & 0.89          & 0.92          & 0.89          & 0.72          & \textbf{0.94}\\
    \hline
    \end{tabular}
    \label{tab:replier_cross_camp_performance}
\end{table}

\section{Discussion}

Understanding the engagement patterns of IO operators is crucial for identifying vulnerable individuals and devising effective countermeasures. 
Our Serbia and Egypt campaign case studies reveal that journalists, state officials, news media, and politicians are primary targets for coordinated IO attacks. 
These attacks can originate either from within the targeted country or from other nation-states.
Our findings further suggest that influential individuals may serve as potential sensors for identifying IO campaigns.

To detect coordinated reply attacks, we propose a campaign-independent and general machine learning framework consisting of a tweet classifier and a replier classifier. 
First, the tweet classifier identifies tweets that receive coordinated replies, narrowing the scope for further investigation. 
This classifier is robust to variations in targeted tweet popularity and general across various campaigns. 
An analysis of the features used by the classifier indicates that the level of engagement received by the replies is the most distinguishing factor. 

Second, the replier classifier identifies operators engaged in coordinated reply attacks. 
In addition to generalizing across campaigns like the tweet classifier, the replier classifier is also capable of handling different levels of class imbalance. 
The most significant features for replier classification are those describing the distribution of similarity between replies from a replier and those from other repliers to the same tweets.

This study presents a proof of concept for the proposed classifiers, evaluated on a number of IO campaigns that have been detected and taken down by Twitter.  
Our experiments were carried out on compute nodes equipped with two 64-core AMD EPYC 7742 2.25 GHz CPUs and 512 GB of RAM. 
The efficacy, efficiency, and robustness of our classifiers would ideally be validated in the wild. 
Unfortunately, due to changes in data-sharing policies from X/Twitter, we are unable to conduct such tests. 

The proposed framework can be extended to other platforms with similar reply functionalities, including Facebook, Threads, Mastodon, and Bluesky. 
Additionally, our models could be developed into products or extensions that users could employ for enhanced online safety.

Our study has potential impacts on platform integrity and public dialogue. 
First, it reveals that what looks like public reactions may in fact be efforts to manipulate a target and other participants of a genuine conversation. 
For instance, politicians may be posting on social media to solicit public opinions. 
Coordinated replies may skew their perception of public sentiment \cite{gerrymandering}. 
Such replies can further distort genuine discourse if portrayed by the media as reflective of public opinion.  
Secondly, coordinated replies enable malicious actors to maximize the public exposure of their posts by exploiting the popularity of their targets, thereby amplifying their influence. 
This pollutes online dialogues with spam, influence campaigns, and divisive messages that harass the targeted individuals or provoke the public.
Such behavior may also alienate the targets and prevent them from sharing their opinion on social media. 
For all these reasons, platforms should protect the targets from coordinated reply attacks.
The research community may use our methodology to detect coordinated reply attacks and study the campaigns, perpetrators, and their potential effects on the individuals. 
Our methodology may also guide the development of countermeasures by social media platforms. 

\paragraph{Ethical Impact.} 

This study has been granted exemption from Institutional Review Board review (
% \anon{
Indiana University protocols  12410 and 1102004860
% }
). Our results can be reproduced using code available at
% \anon{
\url{github.com/osome-iu/io-coordinated-replies}
% } 
and data available at 
% \anon{
\url{doi.org/10.5281/zenodo.13896309}
% }
. 
The collection and release of the dataset comply with the Twitter platform's terms of service. To mitigate the potential ethical risks of analyzing human subjects, we only rely on the data of public Twitter accounts do not include any raw data. We only manually inspect the profiles of the targets of the attacks, who are public figures and constitute the vulnerable group our study aims to protect. We provide our annotation data about these profiles for reproducibility. Our classification models do not use any personally identifiable information. We only report aggregated results. While our main objective is to detect coordinated replies, an attack that is frequently employed by information operations, we acknowledge that it may be also used by regular social media users organizing among themselves for activism. Thus, we suggest that our classifiers should complement human investigation when employed in the wild. Furthermore, they should not be misused to label users as information operation accounts without thorough human verification. 

\paragraph{Acknowledgements.}

% \anon{
This work  was  supported  in  part  by  Knight Foundation,  Craig Newmark Philanthropies, DARPA (contract HR001121C0169), and Lilly Endowment, Inc., through its support for the Indiana University Pervasive Technology Institute.
% }

\bibliography{ref}

\section{Ethics Checklist}

\begin{enumerate}

\item For most authors...
\begin{enumerate}
    \item  Would answering this research question advance science without violating social contracts, such as violating privacy norms, perpetuating unfair profiling, exacerbating the socio-economic divide, or implying disrespect to societies or cultures?
    \answerYes{Yes}
  \item Do your main claims in the abstract and introduction accurately reflect the paper's contributions and scope?
    \answerYes{Yes}
   \item Do you clarify how the proposed methodological approach is appropriate for the claims made? 
    \answerYes{Yes}
   \item Do you clarify what are possible artifacts in the data used, given population-specific distributions?
    \answerYes{Yes}
  \item Did you describe the limitations of your work?
    \answerYes{Yes}
  \item Did you discuss any potential negative societal impacts of your work?
    \answerNA{NA}
      \item Did you discuss any potential misuse of your work?
    \answerNA{NA}
    \item Did you describe steps taken to prevent or mitigate potential negative outcomes of the research, such as data and model documentation, data anonymization, responsible release, access control, and the reproducibility of findings?
    \answerYes{Yes}
  \item Have you read the ethics review guidelines and ensured that your paper conforms to them?
    \answerYes{Yes}
\end{enumerate}

\item Additionally, if your study involves hypotheses testing...
\begin{enumerate}
  \item Did you clearly state the assumptions underlying all theoretical results?
    \answerNA{NA}
  \item Have you provided justifications for all theoretical results?
    \answerNA{NA}
  \item Did you discuss competing hypotheses or theories that might challenge or complement your theoretical results?
    \answerNA{NA}
  \item Have you considered alternative mechanisms or explanations that might account for the same outcomes observed in your study?
    \answerNA{NA}
  \item Did you address potential biases or limitations in your theoretical framework?
    \answerNA{NA}
  \item Have you related your theoretical results to the existing literature in social science?
    \answerNA{NA}
  \item Did you discuss the implications of your theoretical results for policy, practice, or further research in the social science domain?
    \answerNA{Answer}
\end{enumerate}

\item Additionally, if you are including theoretical proofs...
\begin{enumerate}
  \item Did you state the full set of assumptions of all theoretical results?
    \answerNA{NA}
	\item Did you include complete proofs of all theoretical results?
    \answerNA{NA}
\end{enumerate}

\item Additionally, if you ran machine learning experiments...
\begin{enumerate}
  \item Did you include the code, data, and instructions needed to reproduce the main experimental results (either in the supplemental material or as a URL)?
    \answerYes{Yes}
  \item Did you specify all the training details (e.g., data splits, hyperparameters, how they were chosen)?
    \answerYes{Yes}
     \item Did you report error bars (e.g., with respect to the random seed after running experiments multiple times)?
    \answerYes{Yes. Our experiments report averages across 10-fold cross-validation as well as standard errors.}
	\item Did you include the total amount of computation and the type of resources used (e.g., type of GPUs, internal cluster, or cloud provider)?
    \answerYes{Yes}
     \item Do you justify how the proposed evaluation is sufficient and appropriate to the claims made? 
    \answerYes{Yes}
     \item Do you discuss what is ``the cost'' of misclassification and fault (in)tolerance?
    \answerYes{Yes. We recommend manual inspection to complement our classifiers.
    }
  
\end{enumerate}

\item Additionally, if you are using existing assets (e.g., code, data, models) or curating/releasing new assets, \textbf{without compromising anonymity}...
\begin{enumerate}
  \item If your work uses existing assets, did you cite the creators?
    \answerYes{Yes.}
  \item Did you mention the license of the assets?
    \answerNA{NA}
  \item Did you include any new assets in the supplemental material or as a URL?
    \answerYes{Yes. We provide link to data and code as URL.}
  \item Did you discuss whether and how consent was obtained from people whose data you're using/curating?
    \answerYes{Yes; this is addressed by IRB exemption.}
  \item Did you discuss whether the data you are using/curating contains personally identifiable information or offensive content?
    \answerYes{Yes. The Twitter profiles are public information.}
\item If you are curating or releasing new datasets, did you discuss how you intend to make your datasets FAIR?
\answerYes{Yes. The link to data is included in the paper so that anyone can access it, including the details of the meta-data in the Datasheet to make it interoperable and reusable.}
\item If you are curating or releasing new datasets, did you create a Datasheet for the Dataset?
\answerYes{The Datasheet is available at \anon{\url{doi.org/10.5281/zenodo.13682862}}}
\end{enumerate}

\item Additionally, if you used crowdsourcing or conducted research with human subjects, \textbf{without compromising anonymity}...
\begin{enumerate}
  \item Did you include the full text of instructions given to participants and screenshots?
    \answerNA{NA}
  \item Did you describe any potential participant risks, with mentions of Institutional Review Board (IRB) approvals?
    \answerNA{NA}
  \item Did you include the estimated hourly wage paid to participants and the total amount spent on participant compensation?
    \answerNA{NA}
   \item Did you discuss how data is stored, shared, and deidentified?
   \answerNA{NA}
\end{enumerate}

\end{enumerate}

\end{document}